\documentclass[
]{ceurart}

\begin{document}

\copyrightyear{2021}
\copyrightclause{Copyright for this paper by its authors.
  Use permitted under Creative Commons License Attribution 4.0
  International (CC BY 4.0).}

\conference{FIRE'21: Forum for Information Retrieval Evaluation, December 13-17, 2021, India}

\title{Abusive and Threatening Language Detection in Urdu using Boosting based and BERT based models: A Comparative Approach}

\author[1]{Mithun Das}[%
orcid=0000-0003-1442-312X,
email=mithundas@iitkgp.ac.in,
]

\author[1]{Somnath Banerjee}[
orcid=0000-0002-9445-8439,
email=som.iitkgpcse@kgpian.iitkgp.ac.in,
]

\author[1]{Punyajoy Saha}[%
orcid=0000-0002-3952-2514,
email=punyajoys@iitkgp.ac.in,
]

\address[1]{Department of Computer Science and Engineering, Indian Institute of Technology, Kharagpur, West Bengal, India}

\begin{abstract}
  Online hatred is a growing concern on many social media platforms. To address this issue, different social media platforms have introduced moderation policies for such content. They also employ moderators who can check the posts violating moderation policies and take appropriate action. Academicians in the abusive language research domain also perform various studies to detect such content better. Although there is extensive research in abusive language detection in English, there is a lacuna in abusive language detection in low resource languages like Hindi, Urdu etc. In this FIRE 2021 shared task - ``HASOC - Abusive and Threatening language detection in Urdu'' the organisers propose an abusive language detection dataset in Urdu along with threatening language detection.  
  
  In this paper, we explored several machine learning models such as XGboost, LGBM, m-BERT based models for abusive and threatening content detection in Urdu based on the shared task. We observed the Transformer model specifically trained on abusive language dataset in Arabic helps in getting the best performance. Our model came \textbf{First} for both abusive and threatening content detection with an F1score of 0.88 and 0.54, respectively. We have made our code public~\footnote{~\url{https://github.com/hate-alert/UrduAbuseAndThreat}}.

\end{abstract}

\begin{keywords}
  Urdu \sep
  Threat Detection \sep
  Abusive language\sep
  Classification
\end{keywords}

\maketitle

\section{Introduction}

In recent years, social media has become an important means of communication. It has allowed people to share their ideas and opinions instantly. Unfortunately, abusive, threatening, aggressive, etc., languages continue to be used online~\footnote{\url{https://www.ohchr.org/EN/NewsEvents/Pages/sr-minorities-report.aspx}} and endanger the well-being of millions of people. In some cases, it has been reported that online incidents have already turned into crimes against minorities~\cite{hate-speech-websci-19} with some of them leading to severe incidents such as the genocide of the Rohingya community in Myanmar~\footnote{~\url{https://www.bbc.com/news/world-asia-41566561}}, the anti-Muslim mob violence in Sri Lanka~\footnote{\url{https://www.hrw.org/news/2019/07/03/sri-lanka-muslims-face-threats-attacks}}, and the Pittsburg shooting~\footnote{\url{https://en.wikipedia.org/wiki/Pittsburgh_synagogue_shooting}}. Targeted community members may further feel emotional and psychological problems~\cite{vedeler2019hate}. To mitigate the detrimental effectiveness of such posts, the social media platforms, such as Twitter~\footnote{~\url{https://help.twitter.com/en/rules-and-policies/hateful-conduct-policy}} and Facebook~\footnote{~\url{https://transparency.fb.com/bn-in/policies/community-standards/hate-speech/}} have laid down moderation policies and employ moderators~\cite{newton_2019} for maintaining civility in their platforms. However due to huge volume of data streaming in these platforms, it is difficult to screen all posts manually and filter such content.

To this end, several studies have tried to come up with methods to automatically detect abusive content ~\cite{mathew2020hatexplain,Davidson2017AutomatedHS,waseem-etal-2017-understanding,Das2021YouTB}, but very little has been done to identify threats~\cite{alshehri-etal-2020-understanding,9536729}. Further most of the work has been done on English language~\cite{DasNews2020,Amjad2021abuse}, there is a significant lack of resources for abusive and threatening detection in Urdu. Urdu is the fifth most spoken language, with more than 230  million speakers~\footnote{~\url{https://www.statista.com/statistics/266808/the-most-spoken-languages-worldwide/}} around the world. It is the official language of Pakistan. Apart from Pakistan, Urdu is spoken in many countries, including UK, United States, India, and Middle East. Recently, Urdu is also gaining popularity in social media usage~\footnote{\url{https://www.siasat.com/world-Urdu-webinar-Urdu-too-became-language-of-electronic-social-media-1966555/}}. Hence, more effort is required to detect and mitigate such abusive and threatening language in Urdu.

Recently, different shared task like HASOC 2019~\cite{mandl2019overview} have been launched to identify Hate and offensive content in languages other than English. There is also one sub-task in HASOC 2020\footnote{\url{https://hasocfire.github.io/hasoc/2020/index.html}} which aimed to identify offensive post in code-mixed dataset. Extending that task further, the organisers of this shared task~\cite{hasoc2021fireabuse} have build two datasets of 3400, 9950 posts to detect abusive and threatening language in Urdu. Twitter's definition has been followed to describe , whether a post is abusive/non-abusive~\footnote{~\url{https://help.twitter.com/en/rules-and-policies/abusive-behavior}}, and threading/non-threatening~\footnote{~\url{https://help.twitter.com/en/rules-and-policies/glorification-of-violence}}.

In this paper we explore several machine learning models for classification in these sub-tasks. We find that for both the sub-task A \& B, our Transformer based model~\footnote{based on a recent Transformer model dehatebert-arabic \url{https://huggingface.co/Hate-speech-CNERG/dehatebert-mono-arabic}}  ranked \(1st\)  at the shared task. The rest of the paper is organized as follows: In Section 2 we cover some of the related works; We discuss the Dataset Description in Section 3; In Section 4 we present the System Description. Then we discuss about the results, some observation in section 5 and 6. Finally we finish our paper with Conclusion.

\section{Related Work}
Detection of abusive language in natural languages has recently gained significant attraction among the research community. However, the space of abusive language is vast and has its own subtleties.

In 2017, Waseem et al.~\cite{waseem-etal-2017-understanding} classify abusive languages into two category ``Directed" (is the language directed towards a specific individual or entity) and ``Generalized" (directed towards a generalized group), further this category has been divided into another two category ``Explicit" and ``Implicit" (the degree to which it is explicit).

~\citet{Davidson2017AutomatedHS} contributed a dataset in which thousands of tweets were labeled "hate", "offensive", and "neither", with the classification task of detecting hate/offensive speech present in Tweets in mind. Using this dataset, they then explored how linguistic features such as character and word n-grams affected the performance of a classifier aimed to distinguish the three types of Tweet. Additional features in their classification involved binary and count indicators for hashtags, mentions, retweets, and URLs, as well as features for the number of characters, words, and syllables in each tweet. The authors found that one of the issues with their best performing models was that they could not distinguish between hate and offensive posts.

In 2018, ~\citet{Pitsilis2018DetectingOL}, tried recurrent neural networks (RNNs) to identify offensive language in English and found that it was quite effective in this task by achieving 0.9320 F1-score using ensemble methods. RNN’s remember the output of each step the model conducts. This approach can capture linguistic context within a text which is critical to detection. While RNN’s have been projected to work well with language models, other neural network models, such as the CNN. LSTM have had notable success in detecting hate/offensive speech~\cite{Goldberg2015,Sarracn2018HateSD}.

Recently, Transformer based~\cite{Vaswani2017AttentionIA} language models such as, BERT, m-BERT~\cite{devlin2019bert} are becoming quite popular in several downstream task, such as classification, span detection etc. Previously, it has been identified that Transformer based models have been outperformed several deep learning models~\cite{mathew2020hatexplain} such as CNN-GRU, LSTM etc. Having observed the superior performance of these Transformer based model, we focus on building these model for our classification problem.

\section{Dataset Description}

The shared tasks~\cite{hasoc2021overview} present in this competition are divided into two parts. Where in one part participants have to focus on detecting Abusive language using twitter tweets in Urdu language (Subtask A)\footnote{\url{https://ods.ai/competitions/Urdu-hack-soc2021}} and in other part mainly focusing on detecting Threatening language using Twitter tweets in Urdu language (Subtask B)\footnote{\url{https://ods.ai/competitions/Urdu-hack-soc2021-threat}}. The presented data has been collected and annotated from Natural Language and Text Processing Laboratory\footnote{\url{https://nlp.cic.ipn.mx/}} at Center for Computing Research\footnote{\url{https://www.cic.ipn.mx/index.php/en/}} of Instituto Politécnico Nacional, Mexico.

\subsection{Subtask A}

This task is a binary classification task in which tweets need to be classified into two classes, namely: Abusive and Non-Abusive. Training data has total 2400 instances and Testing data has total 1100 instances, which is already annotated as Abusive and Non-Abusive. The mentioned dataset is balanced between Abusive and Non-Abusive classes. The dataset description for Urdu Abusive Task has been represented in Table \ref{tab:binary_abusive}.

\subsection{Subtask B}

This is also a binary classification task of identifying/detecting Threatening language in Urdu. Training  data is having 6000 instances and Testing data has 3950 instances which is annotated as Threatening and Non-Threatening. As per the data statistics it not properly balanced data (ratio is 1:5). We have presented the dataset distribution in Table \ref{tab:binary_threat}.

\begin{table}
\centering
\begin{tabular}{|l|l|l|}
\hline
\multirow{2}{*}{\textbf{Category}} & \multicolumn{2}{l|}{\textbf{Abusive Dataset}} \\ \cline{2-3} 
 & \textbf{Train} & \textbf{Test} \\ \hline
\textbf{Abusive} & 1187 & 563 \\ \hline
\textbf{Non-Abusive} & 1213 & 537 \\ \hline
\textbf{Total} & 2400 & 1100 \\ \hline
\end{tabular}
\caption{Binary Classification on Urdu Abusive Dataset (Subtask A)}
\label{tab:binary_abusive}
\end{table}

\begin{table}
\centering
\begin{tabular}{|l|l|l|}
\hline
\multirow{2}{*}{\textbf{Category}} & \multicolumn{2}{l|}{\textbf{Threatening Dataset}} \\ \cline{2-3} 
 & \textbf{Train} & \textbf{Test} \\ \hline
\textbf{Threatening} & 1071 & 719 \\ \hline
\textbf{Non-Threatening} & 4929 & 3231 \\ \hline
\textbf{Total} & 6000 & 3950 \\ \hline
\end{tabular}
\caption{Binary Classification on Urdu Threatening Dataset (Subtask B)}
\label{tab:binary_threat}
\end{table}

\section{System Description}

In this section, we will explain the details regarding the features and machine learning models used for the task. We have attempted several models for abusive and threatening  detection task. In both of the cases as a baseline we tried XGBoost\cite{Chen_2016} and LightGBM\cite{Ke2017LightGBMAH} with pre-trained Urdu laser embedding. Later we have tried Transformer-based pre-trained architecture of multilingual BERT\cite{devlin2019bert}. The beauty of the mBERT is it is pretrained in unsupervised manner on multilingual corpus. Besides we have used another m-BERT based model which is previously fine-tuned on Arabic hate speech date set. The model has been referred as ``Hate-speech-CNERG/dehatebert-mono-arabic'\cite{aluru2020deep} model. The motivation of using the following model in Arabic language because it is origin of Urdu~\footnote{~\url{https://en.wikipedia.org/wiki/Urdu}}, so further fine-tuning the model with the Urdu dataset may yield better performance.

\subsection{Binary Classification}

Both Subtask A and B is a binary classification problem. We fine tuned BERT Transformer and classifier layer on top and used binary target labels for individual classes. We have used multilingual BERT (mBERT) and dehatebert-mono-arabic for abusive language classification. Binary cross-entropy loss can be computed for this kind of classification task can be mathematically formulated as:

\[Loss = -\sum_{i=1}^{C'=2}t_{i} log (s_{i}) = -t_{1} log(s_{1}) - (1 - t_{1}) log(1 - s_{1})\]

Where it’s assumed that there are two classes: C1 and C2. t1 [0,1] and s1 are the ground truth and the score for C1, and t2=1-t1 and s2=1-s1 are the ground truth and the score for C2. 

\subsection{Weighted Binary Classification}

The main challenge in any classification problem is the imbalance in data. This imbalance in data may create a bias towards the most present labels, which lead to a decrease in classification performance. From Table ~\ref{tab:binary_abusive} and Table ~\ref{tab:binary_threat} we observe that, the dataset for abusive tweet detection is almost balanced, but the threatening detection dataset is highly imbalanced. Lots of research has been done in this domain to make the data balance~\cite{batista2004study}. Oversampling and undersampling are very much popular data balancing methods, but they have coherent disadvantages also. Another method of handling an imbalanced dataset is by using class weight so that the instances which are less in the dataset get more importance while calculating the average loss in the model and we follow this method for threatening tweet detection.

\subsection{Tuning Parameters}
For the classical model such as XGBoost and LGBM, we have used the default setting and trained in on the provided dataset for both tasks.  For Transformer-based models, we have run the models for 10 epochs with Adam optimizer\cite{loshchilov2019decoupled} and initial learning rate of 2e-5.
For the abusive tweet detection, we divided the training data points into 85\% and 15\% split and used the 15\% as a validation set. We predict the test set for the best validation performance. For threat detection, we have used the full dataset for training due to data imbalance and predicted the test set after the completion of 10 epochs.

\begin{table}
\centering
\begin{tabular}{|l|c|c|c|c|} 
\hline
\multicolumn{1}{|c|}{\multirow{2}{*}{Classifiers}} & \multicolumn{2}{l|}{Urdu Abusive Dataset} & \multicolumn{2}{l|}{Urdu Threatening Dataset}  \\ 
\cline{2-5}
\multicolumn{1}{|c|}{}                             & F1 Score         & ROC Curve              & F1 Score         & ROC Curve                   \\ 
\hline
XGBoost                                            & 0.76072          & 0.81962                & 0.24712          & 0.7319                     \\ 
\hline
LGBM                                               & 0.76667          & 0.83937                & 0.20474          & 0.71286                     \\ 
\hline
mBERT                                              & 0.84000          & 0.90221                & 0.46961          & 0.73768                     \\ 
\hline
dehatebert-mono-arabic                                       & \textbf{0.88062} & 0.92449                & \textbf{0.54574} & 0.81096                     \\
\hline
\end{tabular}
\caption{Classification Result on Private Leaderboard (Subtask B)}
\label{tab:threat_public}
\end{table}

\section{Results}
The results of all the models for both sub-tasks are presented in Table ~\ref{tab:threat_public} ( We have shown the classification result of the private leaderboard).  By using mBert based "dehatebert-mono-arabic" model and further fine-tuning it with the Urdu datasets, our model got the first position for both sub-task A \& B, with F1 score of 0.8806 for abusive tweet detection and 0.5457 for threatening tweet detection.

\if{0}
Our observation was among most of the individual Transformer based BERT models, Hate-speech-CNERG/dehatebert-mono-arabic is giving state-of-the-art performance consistently for both of the subtask.\par
While achieving the performance scores we have used multiple random seed, and have observed that performance was heavily getting impacted for different seeds. We have shown the classification result of private leaderboard in Table \ref{tab:threat_public}.
\fi

\section{Discussion}
We found that our model was able to achieve good performance for abusive tweet detection (sub-task A), but was not able to perform well in threatening tweet identification (sub-task B). One of the issues with the dataset of sub-task B is, it is highly imbalanced, and because of that the models are not able to learn the different dimensionality of threatening posts. 

\section{Conclusion}
In this shared task, we have experimented with classical machine learning model as well as Transformer based model. We have used XGboost and LGBM classifier with pre-trained laser embedding. 
We finetuned both mBERT and "dehatebert-mono-arabic" model and observed insteading of finetuning a model from scratch, it is better to use existing finetune model in the same domain. Our model occupied the top position for both sub-task A \& B.

\bibliography{sample-ceur}

\end{document}